\documentclass{article}
\usepackage{spconf,amsmath,graphicx}

\usepackage{enumitem}
\usepackage{multirow}
\setlist{nosep, leftmargin=14pt}

\usepackage{mwe} 

\title{Scale-Aware Curriculum Learning for Data-Efficient Lung Nodule Detection with YOLOv11}
\name{Yi Luo$^{1}$ \qquad Yike Guo$^{1}$ \qquad Hamed Hooshangnejad$^{2}$ \qquad Kai Ding$^{2}$}
\address{$^{1}$Department of Biomedical Engineering, Johns Hopkins University\\
         $^{2}$Department of Radiation Oncology and Molecular Radiation Sciences, Johns Hopkins University}
\begin{document}
%
\maketitle
\begin{abstract}
Lung nodule detection in chest CT is crucial for early lung cancer diagnosis, yet existing deep learning approaches face challenges when deployed in clinical settings with limited annotated data. While curriculum learning has shown promise in improving model training, traditional static curriculum strategies fail in data-scarce scenarios. We propose Scale-Adaptive Curriculum Learning (SACL), a novel training strategy that dynamically adjusts curriculum design based on available data scale. SACL introduces three key mechanisms: (1) adaptive epoch scheduling, (2) hard sample injection, and (3) scale-aware optimization. We evaluate SACL on the LUNA25 dataset using YOLOv11 as the base detector. Experimental results demonstrate that while SACL achieves comparable performance to static curriculum learning on the full dataset in mAP$_{50}$, it shows significant advantages under data-limited conditions with 4.6\%, 3.5\%, and 2.0\% improvements over baseline at 10\%, 20\%, and 50\% of training data respectively. By enabling robust training across varying data scales without architectural modifications, SACL provides a practical solution for healthcare institutions to develop effective lung nodule detection systems despite limited annotation resources.
\end{abstract}
\section{Introduction}
Lung cancer is the leading cause of cancer-related death worldwide \cite{leiter2023global}, with over 238,000 new cases diagnosed annually in the United States \cite{siegel2023cancer}. Despite substantial advances in research and treatment, the overall five-year survival rate remains below 20\% \cite{fitzmaurice2017global}, representing a major public health challenge. Computed tomography (CT) imaging is generally regarded to be the best imaging modality for evaluating lung structure \cite{verschakelen2007computed}, offering exceptional detail in identifying and characterizing pulmonary abnormalities \cite{hoffman2022origins}, This level of detail is crucial for early detection, as the initial stages of lung cancer usually present as small nodules that are round opacity or irregular lung lesions \cite{hansell2008fleischner}.

In recent years, rapid advances in artificial intelligence have spurred the development of computer-aided detection systems for pulmonary nodule analysis. Deep learning approaches for lung nodule detection can be broadly categorized into two-stage and one-stage detectors. Two-stage detectors, exemplified by Faster R-CNN and Mask R-CNN variants adapted for medical imaging \cite{su2021lung,cai2020mask}, first generate region proposals and then perform classification, achieving high sensitivity but at considerable computational cost. In contrast, one-stage detectors from the YOLO family unify detection and classification in a single forward pass, offering faster inference speeds while maintaining competitive accuracy\cite{sapkota2025yolo}. YOLO-based detectors have emerged as a mainstream approach in recent lung nodule detection research, with numerous studies demonstrating their effectiveness in balancing detection accuracy and computational efficiency for medical imaging applications \cite{tang2025lung,song2025improved, liu2025optimized}.

Curriculum learning (CL), inspired by the human learning process where knowledge is acquired progressively from simple to complex concepts, has emerged as a powerful training strategy in deep learning. This approach systematically organizes training samples based on their difficulty levels, allowing neural networks to first master easier examples before tackling more challenging ones \cite{bengio2009curriculum}. In medical imaging analysis, curriculum learning has proven effective in managing the complexity of clinical data, with studies showing that well-designed training curricula can accelerate model convergence and improve detection performance \cite{wang2023grenet,gong2022less,jimenez2022curriculum}.

Despite the promising results of curriculum learning in medical imaging, its application to 3D CT lung nodule detection remains unexplored. While previous lung nodule detection research has primarily relied on the LUNA16 dataset \cite{setio2017validation}, the recently introduced LUNA25 dataset with its significantly expanded scale and diversity presents new opportunities to advance the field with more robust and generalizable models \cite{luna25dataset}.

To address this research gap, we present the first comprehensive study applying curriculum learning to 3D lung nodule detection. Our work makes three key contributions: First, we establish performance baselines for SOTA YOLO detection models on the LUNA25 dataset. Second, we systematically investigate the effectiveness of static CL strategies for 3D lung nodule detection. Third, we propose Scale-Adaptive Curriculum Learning (SACL), which dynamically adjusts curriculum design based on available data scale through adaptive epoch scheduling, hard sample injection, and scale-aware optimization. SACL achieves significant improvements of 4.6\%, 3.5\%, and 2.0\% over baseline methods when using only 10\%, 20\%, and 50\% of training data, respectively, while maintaining comparable performance on the full dataset. This work demonstrates that SACL provides a practical solution for healthcare institutions to develop effective lung nodule detection systems despite limited annotation resources.

\section{Methodology}
\subsection{Dataset}
The LUNA25 grand challenge presents a retrospective, multi-center, comprehensive dataset comprising 4,096 carefully-annotated low-dose chest CT examinations, designed to develop and validate modern AI algorithms for lung nodule malignancy risk estimation. However, the original LUNA25 dataset was specifically designed for malignancy risk evaluation, providing approximate regional annotations rather than precise nodule boundaries, which limits its direct applicability for lung nodule detection tasks.

Recognizing this limitation, Jun et al. have recently developed an enhanced version leveraging the MedSAM2 foundation model for pixel-level segmentation annotations \cite{MedSAM2}. The annotation process followed a systematic two-step approach: (1) automated segmentation using MedSAM2 with point prompts for each identified lesion, and (2) manual refinement by experts, with approximately 880 nodule masks requiring revision to ensure accurate boundary delineation. Based on these precise segmentation masks, we further generated slice-by-slice detection bounding boxes to create a comprehensive resource for lung nodule detection algorithm development and evaluation.

\subsection{Preprocessing}
Our preprocessing pipeline consisted of four stages: (1) lung segmentation using TotalSegmentator \cite{wasserthal2023totalsegmentator} with boundary expansion to ensure complete lung coverage; (2) slice quality assessment based on lung coverage ratio, intensity variance, and structural clarity; (3) nodule-aware slice selection that retained all nodule-containing slices while selecting high-quality background slices at a 1:2 ratio to maintain balanced training data; and (4) nodule filtering to exclude clinically insignificant nodules smaller than 3mm in diameter, followed by CLAHE enhancement for improved contrast. All processed slices were saved as 512×512 PNG images to ensure consistent input dimensions for model training while preserving the aspect ratio through appropriate padding when necessary. This preprocessing pipeline results in 58,999 high-quality slices from 4,069 patients with slice-level bounding box annotations. To ensure robust evaluation and prevent data leakage, we implemented patient-level data splitting, guaranteeing that all slices from a single patient appear exclusively in one subset. This strategy prevents the model from memorizing patient-specific imaging characteristics, leading to more generalizable detection capabilities. The dataset was divided using patient-level 80/10/10 splitting, resulting in 47,086 training slices from 3,255 patients, 6,357 validation slices from 406 patients, and 5,556 test slices from 408 patients. 

\subsection{Scale aware curriculum learning}
\subsubsection{Curriculum Learning}

We first implement a static three-stage curriculum learning approach. Each CT slice is assigned a complexity score $c \in [0,11]$ computed as:
\begin{equation}
  c = f_{\text{cnt}} + f_{\text{size}} + f_{\text{shape}} + f_{\text{qual}},
  \label{eq:complexity}
\end{equation}
where the four factors capture different aspects of detection difficulty: $f_{\text{cnt}}$ (nodule count) assigns 0.5 for nodule-free slices, 1.0 for single nodules, 2.5 for 2--3 nodules, and 4.0 for 4+ nodules; $f_{\text{size}}$ (nodule size) assigns 0.5 when the smallest nodule exceeds 1000 pixels, 1.0 for 400--1000 pixels, and 3.0 for smaller nodules; $f_{\text{shape}}$ (shape irregularity) assigns 0.5 for regular shapes with low aspect-ratio variance, 1.0 for at most one irregular nodule, and 2.0 for multiple irregular cases; and $f_{\text{qual}}$ (image quality) assigns 0.5 for sharp, high-contrast images (Laplacian variance $>$500, contrast $>$30), 1.0 for medium quality, and 2.0 for blurry or low-contrast images. Based on these complexity scores, we organize the training process into three progressive stages, where each stage introduces increasingly complex samples and adjusts the training parameters accordingly. 

\vspace{2pt}\noindent
\textbf{Stage\,1} ({512}{px}): Simple labeled samples + high-quality negative samples, trained for 50 epochs with learning rate $\eta_1 = 0.003$, classification-focused loss weights (box=2.0, cls=4.0, dfl=0.1), minimal augmentation (3° rotation, 5\% translation, 10\% scaling).

\noindent
\textbf{Stage\,2} ({640}{px}): Simple/medium labeled samples +  high/medium quality negatives, 100 epochs, learning rate $\eta_2 = 0.002$, balanced loss weights (box=5.0, cls=2.0, dfl=0.5), moderate augmentation (8° rotation, 10\% translation, 20\% scaling).

\noindent
\textbf{Stage\,3} ({768}{px}): Full training set (all complexity levels + all negative samples), 100 epochs, learning rate $\eta_3 = 0.001$, localization-focused loss weights (box=7.0, cls=1.5, dfl=1.0), strong augmentation (12° rotation, 15\% translation, 30\% scaling).

This three-stage curriculum as illustrated in Figure~\ref{fig:cl_sacl})(a) assumes sufficient annotated data for effective training in each stage. However, lung nodule detection faces inherent data scarcity. Many clinical sites operate with datasets containing only tens or hundreds rather than thousands of annotated CT scans. Under such constraints, the static curriculum becomes poorly calibrated. Early stages risk overfitting with limited data through excessive repetition. This mismatch between the curriculum's design assumptions and clinical data availability motivates an adaptive approach that scales its curriculum design according to the actual training set size.

\subsubsection{Scale-Aware Curriculum Learning}
\begin{figure}[htbp]
    \centering
    \includegraphics[width=0.5\textwidth]{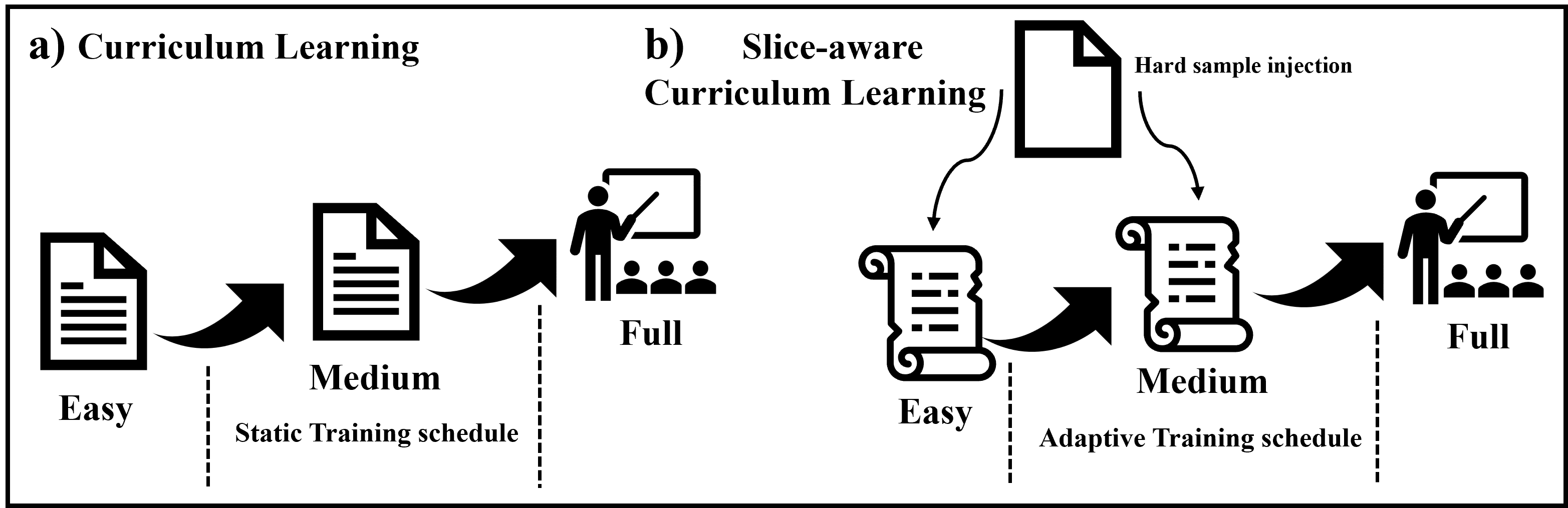}
    \caption{Comparison of CL and SACL training strategies. CL uses fixed stage setups regardless of dataset size, while SACL dynamically adjusts epoch counts, hard sample ratios, and optimization parameters based on available data volume.}
    \label{fig:cl_sacl}
\end{figure}

To address the limitations of static curricula under varying data availability, we propose SACL, as shown in Figure~\ref{fig:cl_sacl}(b).
Given a training subset $D_{\text{sub}} \subseteq D_{\text{full}}$ with relative size $\rho = D_{\text{sub}}/D_{\text{full}}$, SACL dynamically adjusts three key components:

\textbf{(1) Adaptive epoch scheduling.}
The training duration for each stage is scaled according to
\begin{equation}
  E' = \max\{\rho^{\beta} E, \gamma E, E_{\min}\},
\end{equation}
where $E$ is the baseline epoch count, $\beta = 0.7$ controls scaling sensitivity, $\gamma = 0.3$ ensures minimum retention, and $E_{\min} = 20$ prevents degenerate cases.

\textbf{(2) Hard-sample injection.}
To maintain robustness, we enforce a minimum ratio of difficult samples per mini-batch:
\begin{equation}
  r_{\text{hard}}^{\min} = r_0 + (1-\rho)\Delta r,
\end{equation}
with baseline $r_0 = 0.1$ and adjustment factor $\Delta r = 0.3$. This ensures adequate exposure to challenging cases even when data are scarce.

\textbf{(3) Optimization adjustment.}
\begin{equation}
  \eta' = \eta\left[1 - 0.3(1-\rho)\frac{s}{S}\right],
\end{equation}
where $\eta$ is the baseline learning rate, $s$ is the current stage index, and $S$ is the total number of stages. For regularization:
\begin{equation}
  \lambda_{\text{wd}}' = \lambda_{\text{wd}}(2 - \rho) \quad
  p_{\text{drop}}' = \min\{0.3, p_{\text{drop}} + 0.2(1-\rho)\}
\end{equation}
where $\lambda_{\text{wd}}$ is the weight decay coefficient and $p_{\text{drop}}$ is the dropout probability.

Together, these mechanisms enable SACL to maintain effective training dynamics across different data scales, especially for limited scenarios.

\subsection{Experimental Implementation}
All experiments were implemented using YOLOv11, which represents the current SOTA baseline model for lung nodule detection\cite{liu2025optimized}. Training was performed on NVIDIA A100 GPUs with automatic mixed precision enabled for computational efficiency.

\section{Results}
Table~\ref{tab:main_results} presents detection performance across four data scales, revealing distinct patterns in how each training strategy responds to data scarcity.
\begin{table}[h]
\centering
\caption{Performance comparison across different dataset scales. Best results for each dataset size are in bold.}
\vspace{5pt}
\label{tab:main_results}
\begin{tabular}{llcccc}
\hline
Dataset & Method & mAP$_{50}$ & mAP$_{50-95}$ & Recall & Precision \\
\hline
\multirow{3}{*}{100\%} & Baseline & 67.22 & 36.87 & 58.75 & 72.52 \\
 & CL & \textbf{69.37} & \textbf{38.70} & 61.99 & \textbf{75.84} \\
 & SACL & 69.06 & 35.11 & \textbf{63.10} & 71.39\\
\hline
\multirow{3}{*}{50\%} & Baseline & 64.20 & 35.10 & 55.55 & 69.55 \\
 & CL & 63.65 & 34.30 & 56.50 & 68.20 \\
 & SACL & \textbf{65.50} & \textbf{34.60} & \textbf{57.60} & \textbf{71.10} \\
\hline
\multirow{3}{*}{20\%} & Baseline & 57.44 & 29.86 & 49.38 & 64.44 \\
 & CL & 58.16 & \textbf{30.56} & 49.08 & 64.34 \\
 & SACL & \textbf{59.46} & 30.54 & \textbf{50.38} & \textbf{65.46} \\
\hline
\multirow{3}{*}{10\%} & Baseline & 53.18 & 26.95 & \textbf{46.26} & 58.89 \\
 & CL & 54.00 & 27.95 & 43.77 & 62.05 \\
 & SACL & \textbf{55.61} & \textbf{28.73} & 44.71 & \textbf{61.62} \\
\hline
\end{tabular}
\end{table}

CL attains the highest mAP$_{50}$ (69.37\%) and mAP$_{50\text{-}95}$ (38.70\%), confirming the benefit of a well–designed fixed curriculum when ample data are available. SACL reaches a comparable mAP$_{50}$ (69.06\%) and yields the best recall (63.10\%).  Full dataset deactivates most scale factors, SACL and CL differ only by the guaranteed hard–sample floor; the additional hard cases raise recall but slightly hurt the stricter mAP$_{50\text{-}95}$.
  
With 50\% of the training data, CL underperforms the baseline by 0.55pp in mAP$_{50}$. SACL adapts its curriculum parameters to the reduced data volume, achieving 65.50\% mAP$_{50}$ and surpassing both baseline and CL across all metrics. At the 20\% data level, baseline, CL, and SACL achieve mAP$_{50}$ values of 57.44\%, 58.16\%, and 59.46\%, respectively. While CL yields the highest mAP$_{50\text{-}95}$, its improvement over baseline is limited. SACL outperforms both alternatives with +2.02pp in mAP$_{50}$ and leads in recall and precision, demonstrating that its scale-aware adjustments effectively balance learning from hard samples and preventing overfitting under moderate data scarcity, whereas the fixed CL schedule provides only marginal benefits. At the extreme setting of 10\% data, mAP$_{50}$ drops to 53.18\% for the baseline and 54.00\% for CL. However, SACL maintains 55.61\% (+2.43pp over baseline) and achieves the highest mAP$_{50\text{-}95}$ (28.73\%).  
 
The consistent superiority of SACL at this scale confirms that its adaptive control of stage length, sample difficulty, and regularisation continues to deliver robust gains when annotations are extremely limited.
\begin{figure}[htbp]
    \centering
    \includegraphics[width=0.5\textwidth]{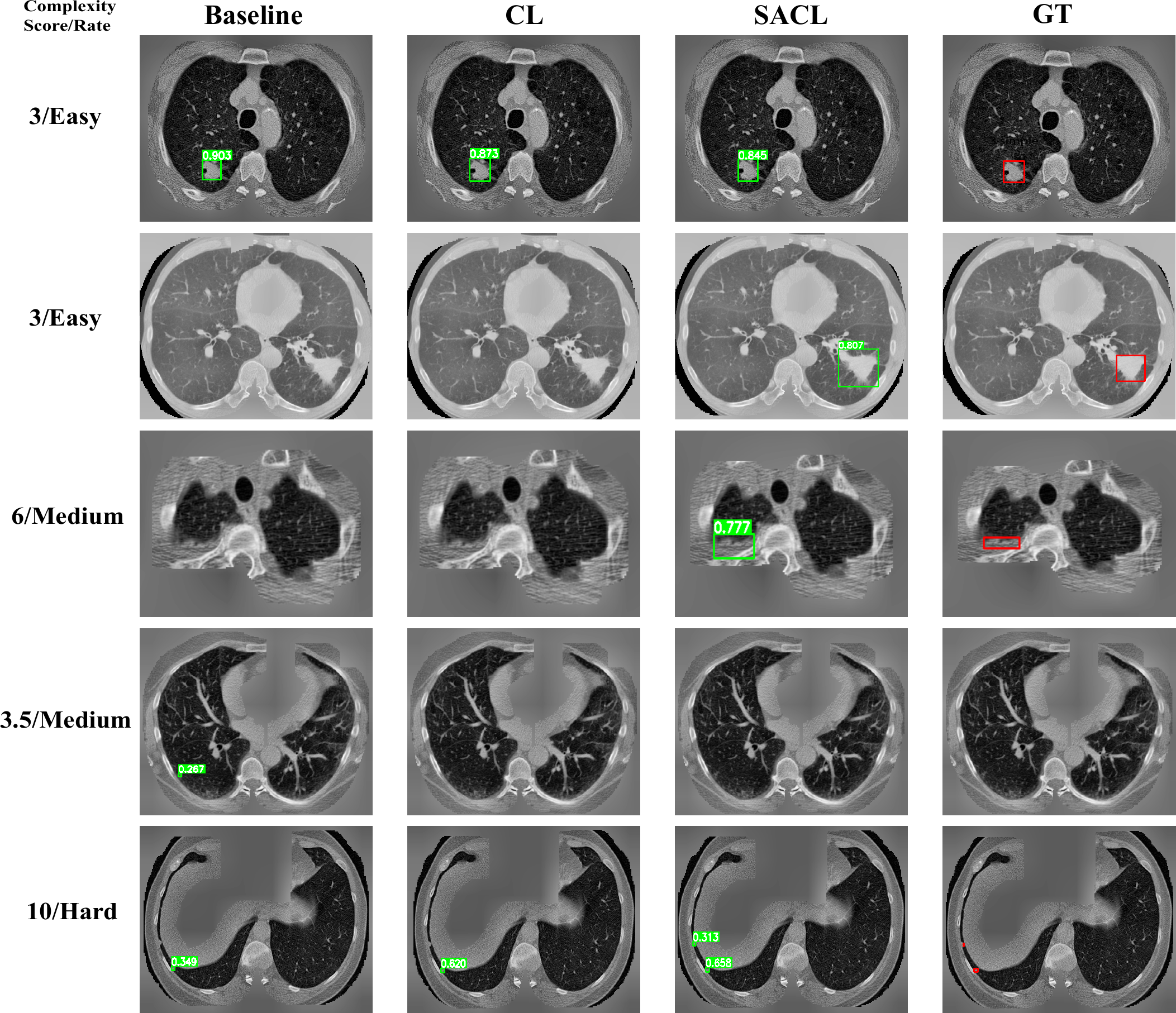}
    \caption{Detection results comparison using 10\% of training data. The first column shows the complexity score/rate for each case. The subsequent columns show results from different models: baseline, CL, SACL, and ground truth. Each row represents the same case. Green boxes indicate model predictions with confidence scores, red boxes show ground truth annotations. Absence of red boxes indicates negative samples, while absence of green boxes indicates the model predicted negative.}
    \label{fig:case_study}
\end{figure}

As illustrated in Figure~\ref{fig:case_study}, the visualization reveals distinct performance patterns across the three models on cases with varying complexity scores/rates as defined in Equation~\ref{eq:complexity}. In the first case (3, Easy), all three models successfully detected the nodule. In the second case (3, Easy) and the third case (6, Medium), only SACL correctly identified the nodules, while both the baseline and CL models failed to detect the ground truth nodules, resulting in false negatives. The fourth case represents a true negative sample where both CL and SACL correctly predicted the absence of nodules, whereas the baseline model erroneously generated a false positive detection. In the final and most challenging case (10, Hard), which contains two nodules, the baseline and CL models detected only the larger nodule, while the SACL model successfully identified both nodules, demonstrating superior performance.

\section{Discussion}
Our results demonstrate SACL's effectiveness in addressing data scarcity in medical imaging. While evaluated on LUNA25, the approach shows promise for other domains including chest X-ray datasets (CheXpert, MIMIC-CXR) and diverse tasks like pneumonia detection and organ segmentation. However, some challenging samples identified by our method may actually hinder model training rather than improve it, necessitating further manual review to distinguish genuinely informative hard samples from potentially mislabeled or ambiguous cases. Additionally, real-world deployment requires validation through prospective studies and robustness testing across different scanners and populations.

The current SACL design employs complexity scoring based on nodule count, size, shape, and image quality to construct curriculum stages. Future improvements could explore dynamic curriculum boundaries that adapt during training based on model performance, rather than fixed complexity thresholds. Additionally, incorporating patient-level difficulty factors such as comorbidities or anatomical variations could create more clinically meaningful curriculum designs. The curriculum could also benefit from multi-modal integration, where difficulty assessment combines imaging features with clinical context, such as patient history and lab results, to better reflect the holistic diagnostic process radiologists employ. This would enable training strategies that mirror real clinical decision-making workflows.

\section{Conclusion}

We presented a novel training strategy SACL that dynamically adapts curriculum learning based on available data scale. Our evaluation on lung nodule detection using the LUNA25 dataset demonstrates that SACL achieves comparable performance to static curriculum learning on the full dataset in mAP50, while showing significant advantages under data-limited conditions. The key innovation of SACL lies in its three adaptive mechanisms: adaptive epoch scheduling, hard sample injection, and scale-aware optimization, which dynamically adjust training parameters based on data availability to overcome the limitations of static curriculum strategies on smaller datasets. By enabling robust training across varying data scales, SACL empowers healthcare institutions with limited annotation resources to develop effective AI-assisted lung nodule detection systems, providing a solution for real-world clinical deployment where data availability remains a persistent challenge.
\bibliographystyle{IEEEbib}
\bibliography{strings,refs}

\end{document}